\documentclass[conference]{IEEEtran}
\IEEEoverridecommandlockouts
\usepackage{cite}
\usepackage{amsmath,amssymb,amsfonts}
\usepackage[ruled]{algorithm2e}
\usepackage{graphicx}
\usepackage{textcomp}
\usepackage[dvipsnames]{xcolor}
\usepackage{colortbl}
\usepackage{multirow}
\usepackage{booktabs}
\usepackage{url}
\usepackage[a4paper, total={184mm,239mm}]{geometry}
\usepackage[hang,flushmargin]{footmisc}
\makeatletter
\newcommand{\algorithmfootnote}[2][\footnotesize]{%
  \let\old@algocf@finish\@algocf@finish
  \def\@algocf@finish{\old@algocf@finish
    \leavevmode\rlap{\begin{minipage}{\linewidth}
    #1#2
    \end{minipage}}%
  }%
}
\makeatother

\definecolor{commentcolor}{RGB}{110,154,155}   

\newcommand{\PyComment}[1]{\ttfamily\textcolor{commentcolor}{\# #1}}  

\newcommand{\PyCode}[1]{\ttfamily\textcolor{black}{#1}} 

\def\BibTeX{{\rm B\kern-.05em{\sc i\kern-.025em b}\kern-.08em
    T\kern-.1667em\lower.7ex\hbox{E}\kern-.125emX}}
\begin{document}

\title{Cocktail: Chunk-Adaptive Mixed-Precision Quantization for Long-Context LLM Inference
}
\author{\IEEEauthorblockN{Anonymous Authors}}
\author{\IEEEauthorblockN{
    Wei Tao$^{\spadesuit}$$^{\heartsuit}$,
    Bin Zhang$^{\spadesuit}$$^{\heartsuit}$,
    Xiaoyang Qu$^{\heartsuit}$,
    Jiguang Wan$^{\spadesuit*}$\thanks{$^{*}$Jigaung Wan (email: jgwan@hust.edu.cn) and Jianzong Wang (email: jzwang@188.com) are the corresponding authors.},
    Jianzong Wang$^{\heartsuit*}$}
    \IEEEauthorblockA{$^{\spadesuit}$Huazhong University of Science and Technology, Wuhan, China}
    \IEEEauthorblockA{$^{\heartsuit}$Ping An Technology (Shenzhen) Co., Ltd, Shenzhen, China}
}
\maketitle

\begin{abstract}
Recently, large language models (LLMs) have been able to handle longer and longer contexts. However, a context that is too long may cause intolerant inference latency and GPU memory usage. Existing methods propose mixed-precision quantization to the key-value (KV) cache in LLMs based on token granularity, which is time-consuming in the search process and hardware inefficient during computation. This paper introduces a novel approach called Cocktail, which employs chunk-adaptive mixed-precision quantization to optimize the KV cache. Cocktail consists of two modules: chunk-level quantization search and chunk-level KV cache computation. Chunk-level quantization search determines the optimal bitwidth configuration of the KV cache chunks quickly based on the similarity scores between the corresponding context chunks and the query, maintaining the model accuracy. Furthermore, chunk-level KV cache computation reorders the KV cache chunks before quantization, avoiding the hardware inefficiency caused by mixed-precision quantization in inference computation. Extensive experiments demonstrate that Cocktail outperforms state-of-the-art KV cache quantization methods on various models and datasets.
\end{abstract}


\section{Introduction}
Large language models have found widespread applications across various domains \cite{10546707, yuan2023large, sun2023test,xiao2024llm}. To enhance their expressive capabilities, the context length of these models has been steadily increasing \cite{chen2023extending,peng2023yarn}. However, extended context also introduces significant challenges. Specifically, longer context lengths lead to a notable increase in inference latency and a substantial rise in memory consumption. The primary reason for the excessive inference latency and high memory consumption in LLMs with long contexts is the KV cache. For instance, in the case of the Llama2 13B model, a context with a length of 128K requires around 100GB of memory to store the KV cache, whereas a single NVIDIA A100 GPU has only 80GB of memory. Besides, during inference, the KV cache must frequently be loaded from the GPU to the GPU's high-speed cache, which is a significant contributor to the high inference latency. 

\begin{figure}[ht]
    \centering
    \includegraphics[width=0.48\textwidth]{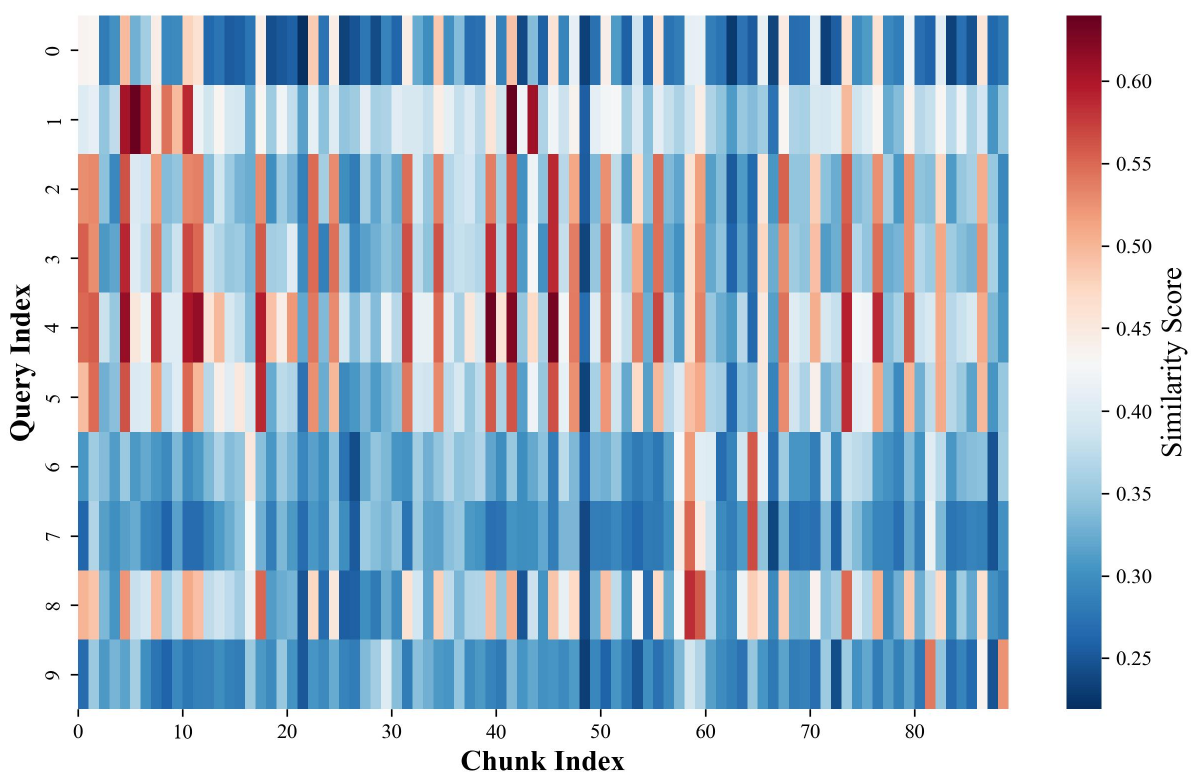}
    \caption{The similarity heatmap between a long passage and 10 different queries. Most of the passage chunks are irrelevant to the query.}
    \label{fig:similar}
    \vspace{-3mm}
\end{figure}

Various methods have been proposed to compress the KV cache, such as attention pruning \cite{shazeer2019fast,ainslie2023gqa,cai2024medusa}, removing or quantizing unimportant tokens \cite{zhang2024h2o, xiao2023efficient,han2023lm}, and considering optimizations from a system perspective \cite{dao2022flashattention, kwon2023efficient,dao2023flashattention}. Among them, the simplest and most effective approach is quantization. Quantization means converting the KV cache from the original FP16 type to a lower bitwidth type (such as INT8, INT4, or even INT2). However, most existing KV cache quantization efforts \cite{liu2024kivi, sheng2023flexgen, zhao2024atom, lin2024qserve} uniformly quantize the KV cache to a single bitwidth. In reality, the importance of tokens within the KV cache is hierarchical, meaning some tokens are more important than others. Therefore, uniform quantization can easily cause accuracy degradation. Some papers \cite{kim2023squeezellm,yang2024no,hooper2024kvquant,dong2024qaq} have proposed implementing mixed-precision quantization based on the importance of the tokens to address this issue, but their quantization search strategies are token-level and very time-consuming. Besides, mixed-precision quantization can cause hardware inefficiency during the inference computation.

In this paper, we propose to apply the mixed-precision quantization to the KV cache at a larger granularity, i.e., the chunk level. We select a long passage (which can be split into 89 chunks) to simulate a long context and create 10 different queries, calculating the similarity between them. As shown in Figure \ref{fig:similar}, for each query, there is only a small portion of chunks that are highly relevant, while most are irrelevant. Obviously, for the KV cache corresponding to context chunks that are highly relevant to the query, we need to retain their precision; otherwise, the model would lose critical information. For the KV cache corresponding to context chunks that are irrelevant to the query, quantizing them to very low bits will not significantly affect the model's accuracy. Based on these observations, we propose a chunk-adaptive mixed-precision KV cache quantization method called Cocktail.

Cocktail mainly contains two modules: chunk-level quantization search and chunk-level KV cache computation. Chunk-level quantization search calculates the similarity scores between the query and each context chunk and determines the bitwidth configuration of the KV cache chunks corresponding to the context chunks according to those scores.
This module maintains the accuracy of LLMs quickly and effectively. We further design the chunk-level KV cache computation module to avoid hardware inefficiency. Chunk-level KV cache computation reorders the context KV cache chunks to ensure the same bitwidth configuration in which KV cache chunks are arranged contiguously in physical memory. This module significantly optimizes the inference latency and memory usage during LLM inference.

In summary, our contributions can be outlined as follows:
\begin{itemize}
\item We propose a chunk-adaptive mixed-precision KV cache quantization method called Cocktail for long-context LLM inference.
    \item Cocktail contains the chunk-level quantization search module, which determines the bitwidth configuration of KV cache chunks based on the similarity scores between the corresponding context chunks and the query, maintaining the inference accuracy of the LLM.
    \item Cocktail contains the chunk-level KV cache computation module, which reorders KV cache chunks in different bitwidth configurations before quantization, avoiding the issue of hardware inefficiency during inference caused by mixed-precision quantization.
    \item Extensive experiments validate that Cocktail achieves better model accuracy, lower latency, and reduced memory usage compared to state-of-the-art (SOTA) methods.
\end{itemize}

\section{Background}
\subsection{LLM Inference}
The computation in LLMs inference primarily revolves around the attention module, with the core focus being the calculation of the $Q, K, V$ matrices \cite{vaswani2017attention}. LLM inference consists of a prefill phase and several decode phases. During the prefill phase, the LLM processes the input tokens (i.e., the concatenation of context and query) to generate the first output token. Subsequently, in each decode phase, the LLM processes the input token along with all previously generated output tokens to produce the next token. Due to the feature of attention computation, the LLM needs to compute the K  and V matrics corresponding to the input tokens and all previously generated output tokens, which is very time-consuming. Therefore, later researchers proposed caching the generated K  and V matrices after processing the input tokens and the previously output tokens, known as KV Cache, so that they do not need to be recomputed in subsequent decode phases. However, as the length of the context increases \cite{chen2023extending,peng2023yarn}, the number of input tokens grows, and the KV cache also becomes larger, leading to severe memory and latency issues.
\subsection{KV Cache Compression}
Past researchers have proposed various methods to compress the KV Cache. Some have suggested pruning the attention module within the KV Cache \cite{shazeer2019fast,ainslie2023gqa,cai2024medusa}. For example,  Shazeer \cite{shazeer2019fast} proposed MQA, where all attention heads share a single set of KV parameters. Some have proposed evicting unimportant tokens from the KV cache \cite{xiao2023efficient,zhang2024h2o,han2023lm}. For instance, Xiao et al. \cite{xiao2023efficient} introduced streamingLLM, which retains only the most recent sliding window of KV Cache along with the initial tokens. Zhang et al. \cite{zhang2024h2o} proposed Q-Hitter, which evaluates the importance of attention based on the sum of attention scores across each token's column in the attention matrix. Others have approached the problem from a system perspective \cite{dao2022flashattention, kwon2023efficient,dao2023flashattention}. For example, Kwon et al. \cite{kwon2023efficient} introduced PageAttention, which applies the memory paging concept from traditional operating systems by mapping logically contiguous KV Cache pages to physically non-contiguous pages through a page table. Dao et al. \cite{dao2022flashattention} proposed FlashAttention, which rewrites certain complex operators to be processed as much as possible within the GPU's high-speed cache. However, these methods are not as easy to implement as quantitative methods, and their optimization results are also not as effective as those achieved by quantitative methods.

\subsection{LLM Quantization}
The quantization work for LLMs initially focused on quantizing weights and activations \cite{xiao2023smoothquant, frantar2022gptq,dettmers2022gpt3,ge2023model,dettmers2023spqr,lin2024awq}. For example, Xiao et al. \cite{xiao2023smoothquant} proposed SmoothQuant, which scales weights that are easy to quantize and activations that are difficult to quantize separately.
However, during inference with long contexts, the KV cache often becomes larger than the weights and activations. Therefore, the quantization work for LLMs has been extended to the KV cache as well \cite{sheng2023flexgen,zhao2024atom,liu2024kivi,lin2024qserve}. For instance, Zhao et al. \cite{zhao2024atom} introduced Atom, which performs group quantization of the KV cache to low bits, where each group has independent quantization parameters. Liu et al. \cite{liu2024kivi} proposed KIVI, which applies per-channel quantization to the K cache while keeping the original per-token quantization for the V cache. However, these quantization methods that uniformly quantize tokens to the same bitwidth cannot maintain model accuracy effectively. Other researchers have proposed mixed-precision quantization methods to process important tokens in LLMs \cite{kim2023squeezellm,yang2024no,hooper2024kvquant,dong2024qaq}. For example, Kim et al. \cite{kim2023squeezellm} proposed SqueezeLLM, which divides the weights into dense matrices without outliers and sparse matrices containing outliers, then applies low-bit quantization to the dense matrices while keeping the outliers in FP16 precision. Yang et al. \cite{yang2024no} proposed MiKV, which identifies unimportant tokens based on attention scores but quantizes them instead of evicting them. Hooper et al. 
 \cite{hooper2024kvquant} introduced KVQuant, which uses a custom data type called nuqX to represent the mixed-precision quantized KV Cache. These methods use token-level quantization search, which is very time-consuming, and they have not adequately addressed the hardware inefficiency issues brought by mixed-precision quantization.

\begin{figure*}[ht]
    \centering
    \includegraphics[width=0.95\textwidth]{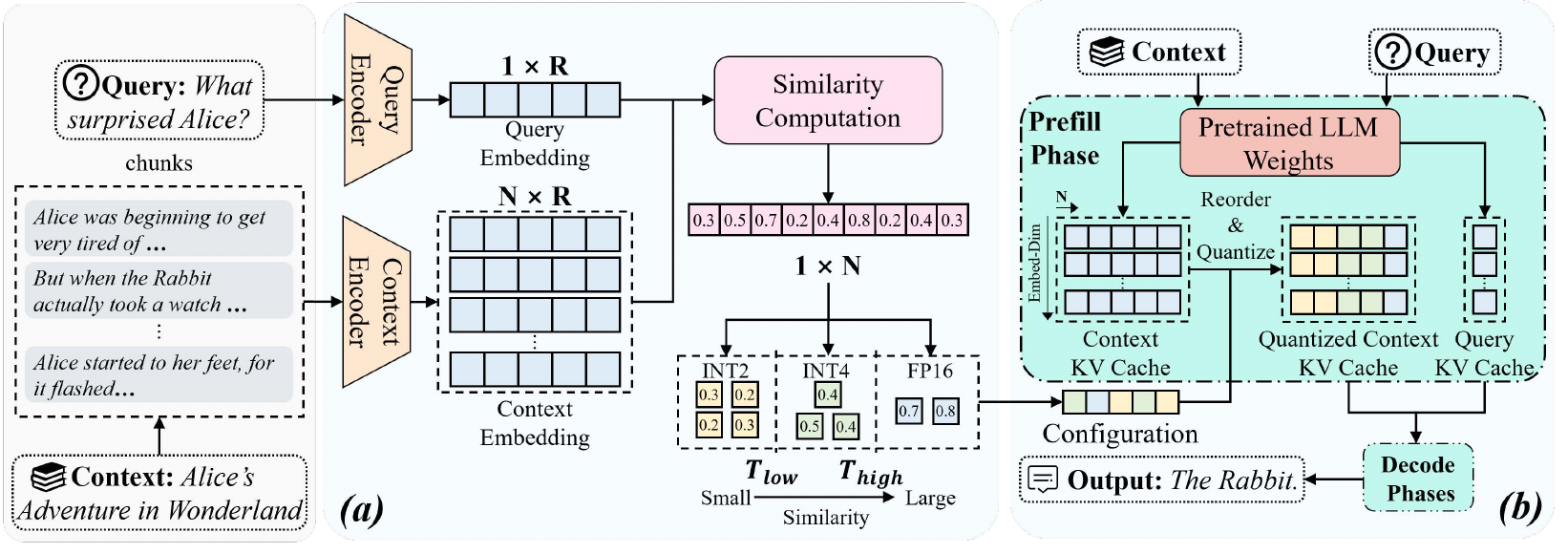}
    \caption{The architecture overview of Cocktail. (a) The chunk-level quantization search module. (b) The chunk-level KV cache computation module.}
    \vspace{-3mm}
    \label{fig:system}
    
\end{figure*}
\section{Method}
\subsection{Architecture Overview}
The architecture of Cocktail is shown in Figure \ref{fig:system}. The long context is first segmented into several equal-length, short chunks (If the length of the context is not divisible by the chunk size, we truncate the portion at the end of the context that cannot be divided by the chunk size. The KV cache of this portion will be kept in FP16 precision).
Then, the chunk-level quantization search receives and processes the context chunks and the query, outputting the quantization bitwidth configuration. Next, the context chunks and the query are fed into the pre-trained LLM for formal inference. During the prefill phase, we reorder the KV cache chunks generated from the context chunks. The reordered KV cache chunks output from this module are subsequently quantized according to the configuration provided by the chunk-level quantization search module. Since the quantized context KV cache in Figure \ref{fig:system} resembles a horizontal three-layered cocktail with distinct colors, we name our method \textit{Cocktail}. Finally, after several decode phases, the pre-trained LLM outputs the answer. To maintain model accuracy, we only quantize the KV cache of the context while retaining FP16 precision for the KV cache of the output tokens generated in the decode phases. Given that, in long-context LLM inference datasets, the length of the context is significantly larger than that of the output, we believe this strategy will not result in significant memory or inference latency overhead.

\subsection{Chunk-level Quantization Search}
Chunk-level quantization search draws inspiration from the concept of RAG (Retrieval-Augmented Generation) \cite{lewis2020retrieval}. RAG retrieves documents similar to the input query from a third-party corpus. These retrieved documents are concatenated with the query to improve the model’s understanding. To retrieve the appropriate documents, RAG employs an encoder to encode both the query and the candidate documents, calculates the similarity between the encoded vectors, and selects the top-k documents with the highest similarity. In our chunk-level quantization search, we use the encoder from RAG to distinguish the similarity between different context chunks and the query so that we can quickly determine the quantization bitwidth configuration for their corresponding KV cache chunks. Specifically, we use the Facebook-Contriever \cite{izacard2021unsupervised} model as both the context encoder and the query encoder. 

As shown in Figure \ref{fig:system}(a), context chunks and query are sent separately to the chunk encoder and query encoder. After encoding, we obtain the context chunk embeddings and query embedding. Next, we calculate the cosine similarity between the query embedding and each context chunk embedding, resulting in a similarity score list. The cosine similarity formula is:
\begin{equation}
    sim(q, c_i) = \frac{q \cdot c_i}{||q|| \times ||c_i||} \quad i = 1,2,...,N
\end{equation}
where $q$ is the query embedding, $c_i$ is the $i_{th}$ chunk embedding, $N$ is the number of chunk embeddings. $||\cdot||$ means the L2 norm.

Subsequently, we set two thresholds, $T_{low}$ and $T_{high}$ ($0$ \textless\ $T_{low}$ \textless\ $T_{high}$ \textless $1$). We compare the similarity score at each index in the similarity score list with these two thresholds. For a similarity score greater than $T_{high}$, the context chunk at that index is considered highly relevant to the query, and we set the bitwidth configuration of its corresponding KV cache chunk as FP16. For a similarity score less than $T_{low}$, the context chunk at that index is considered to have little relevance to the query, and we set the bitwidth configuration of its corresponding KV cache chunk as INT2. For a similarity score between $T_{low}$ and $T_{high}$, we adopt a compromise strategy, setting the bitwidth configuration of its corresponding KV cache chunk as INT4. We use two hyperparameters $\alpha$ and $\beta$ to control the values of these two thresholds:
\begin{equation}
    T_{low} = s_{min} + (s_{max} - s_{min}) \times \alpha
\end{equation}
\begin{equation}
    T_{high} = s_{max} - (s_{max} - s_{min}) \times \beta
\end{equation}
where $s_{min}$ and $s_{max}$ are the minimal and maximal values in the similarity score list, respectively. We will discuss the influence of different $\alpha$ and $\beta$ on the model performance in Section \ref{subsec:analysis}.

\begin{figure}
    \centering
    \includegraphics[width=0.48\textwidth]{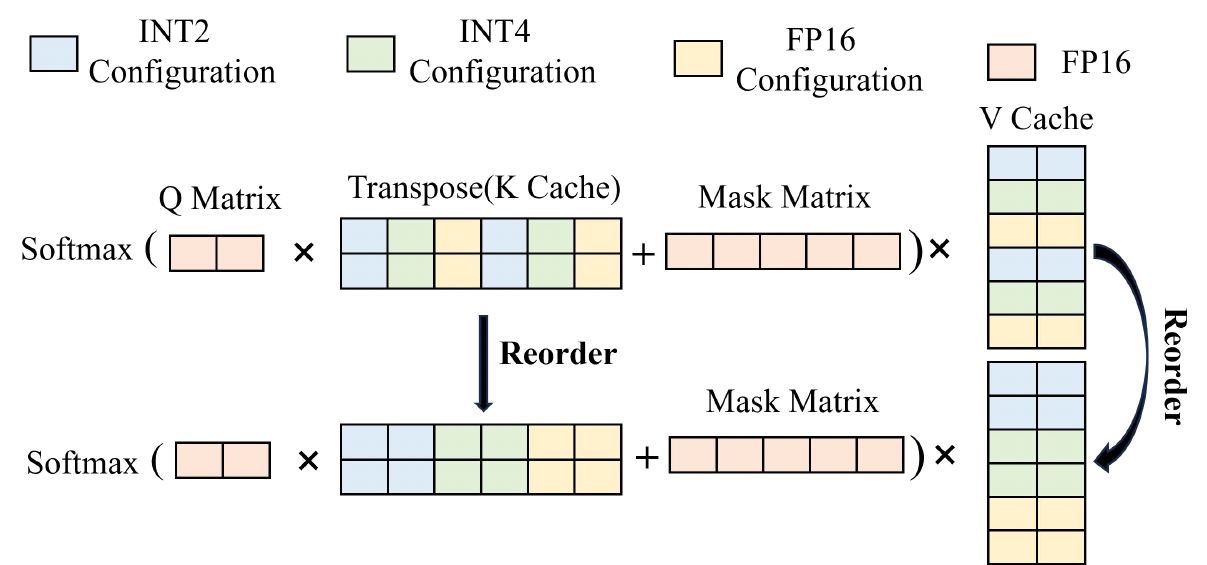}
    \caption{The process of KV cache chunk reordering.}
    \label{fig:reorder}
    \vspace{-6mm}
\end{figure}
\subsection{Chunk-level KV Cache Computation}
Directly applying mixed-precision quantization can result in chunks with different bitwidth that are physically contiguous, which can lead to hardware inefficiency during inference computation. For example, modern GPUs use cache lines to optimize data reading, but data with different bitwidths cannot be aligned properly, potentially spanning multiple cache lines. This requires loading more cache lines, increasing the cache miss rate and memory access frequency, which significantly raises inference latency. Besides, if the hardware is designed to read multi-byte data simultaneously, such as single instruction multiple data (SIMD) architecture, some of the narrower bitwidth data may only use part of the hardware resources, leading to GPU memory waste.

Therefore, we design the chunk-level KV cache computation module to address the hardware inefficiency problem. The algorithm pseudocode is shown in Algorithm \ref{alg:1}. Specifically, during the prefill phase, we first adopt KV cache chunk reordering. As shown in Figure \ref{fig:reorder}, the KV cache chunks with the same bitwidth configuration are arranged together, making them contiguous in physical memory. Next, these chunks are quantized following the bitwidth configuration determined by the KV cache quantization search module. During the decode phase, we multiply the $Q$ matrix with the transposed three blocks of the $K$ matrix with three different bitwidths to obtain three attention matrix blocks. These blocks are then concatenated along the last dimension to form the attention matrix. The attention matrix is then added to the mask matrix and processed with the softmax function. The processed attention matrix is divided into three blocks again, and each is multiplied by the $V$ matrix blocks with the corresponding bitwidth to produce three small output matrices. The final output matrix is obtained by summing these three small output matrices.

\begin{algorithm}[ht]
\small
\caption{Pseudocode of Chunk-level KV Cache Computation in a Pytorch-like Style}
\algorithmfootnote{quant: the quantization function; fqm: FP16 matrix and quantized matrix multiply; mm: FP16 matrix multiply; cat: concatenation.}
\label{alg:1}

    \PyComment{s: the similarity score list}  \\
    \PyComment{N: the number of context chunks} \\
    \PyComment{K, V: the K, V cache of the context} \\
    \PyComment{T\_low, T\_high: the two thresholds} \\
    
    \PyComment{\textbf{During the prefill phase:}} \\
    \PyCode{for i in range(N):} \\
    \Indp   
        
        \PyCode{if s[i] < T\_low:} \\
        \Indp
            \PyCode{K\_int2.append(K[i])}
            \\ 
            \PyCode{V\_int2.append(V[i])}
            \\ 
        \Indm
        \PyCode{elif s[i] > T\_high:} \\
        \Indp
            \PyCode{K\_fp16.append(K[i])}
            \\ 
            \PyCode{V\_fp16.append(V[i])}
            \\ 
        \Indm
        \PyCode{else:} \\
        \Indp
            \PyCode{K\_int4.append(K[i])}
            \\ 
            \PyCode{V\_int4.append(V[i])}
            \\ 
        \Indm
    \Indm 
    K\_q2, K\_q4 = quant(K\_int2), quant(K\_int4) \\
    V\_q2, V\_q4 =  quant(V\_int2), quant(V\_int4) \\
    \PyComment{Q: the Q vector of the current token} \\
    \PyComment{mask: attention mask matrix} \\
    \PyComment{len\_2, len\_4 = len(K\_q2), len(K\_q4)} \\
    \PyComment{\textbf{During the decode phase:}} \\
    \PyCode{att = fqm(Q,transpose(K\_q2),2)} \\
    \PyCode{att = cat(att,fqm(Q,transpose(K\_q4),4),-1)} \\
    \PyCode{att = cat(att,mm(Q,transpose(K\_fp16)),-1)} \\
    \PyCode{att = softmax(att + mask)} \\
    \PyCode{output = fqm(att[:len\_2)],V\_q2)} \\
     \PyCode{output += fqm(att[len\_2:len\_2+len\_4],V\_q4)} \\
     \PyCode{output += mm(att[len\_2+len\_4:],V\_fp16)} \\
     
\end{algorithm}

We prove that the output obtained in this way is the same as the output from the traditional computation method:
In traditional computation method, assuming there are $N$ context chunks, then $K$ matrix can be divided into a block matrix of the form $[K_1|K_2|...|K_N]^T$, and $V$ matrix can be divided into a block matrix of the form $[V_1|V_2|...|V_N]^T$, both of which are partitioned along the token length dimension. Obviously, attention matrix $A$ can also be divided into $[A_1|A_2|...|A_N]$. According to block matrix multiplication, the final output matrix $O$ is:
\begin{equation}
    O = [A_1V_1 + A_2V_2 + ... + A_NV_N]
\end{equation}

Now, if we perform KV cache chunk reordering, it is equivalent to changing the order of the sub-blocks within the $K$ and $V$ matrices. Suppose after reordering, $K$ becomes 
$[K_{x_1}|K_{x_2}|...|K_{x_N}]^T$, and $V$ becomes $[V_{x_1}|V_{x_2}|...|V_{x_N}]^T$, where $\{x_1, x_2, ..., x_N\}$ is a permutation of $\{1, 2, ..., N\}$. Since the softmax function is not affected by the order of the matrix sub-blocks, the sub-blocks of matrix $A$ will eventually be arranged in the same order as the sub-blocks within $K$, which means $A$ is $[A_{x_1}|A_{x_2}|...|A_{x_N}]$. The final output $O'$ will be:
\begin{equation}
   O' =  [A_{x_1}V_{x_1} + A_{x_2}V_{x_2} + ... + A_{x_N}V_{x_N}]
\end{equation}
which is obviously equal to the matrix $O$ due to the commutative invariance of matrix addition. 
\begin{table}[htbp]
\centering
\caption{Evaluation dataset and metrics.}
\label{tab:dataset}
\begin{tabular}{lll}
\toprule
\textbf{Dataset} & \textbf{Task}  & \textbf{Evaluation Metric} \\ \midrule
Qasper        & Single-Document QA & F1-score                   \\
QMSum         & Summarization      & ROUGE score                \\
MultiNews     & Summarization      & ROUGE score                \\
TREC          & Few-shot Learning  & Classification score       \\
TriviaQA      & Few-shot Learning  & F1 score                   \\
SAMSum        & Few-shot Learning  & ROUGE score                \\
LCC           & Code Completion    & Similarity score           \\
RepoBench-P   & Code Completion    & Similarity score           \\ \bottomrule
\end{tabular}
\vspace{-3mm}
\end{table}

\section{Evaluation}
\begin{table*}[ht]
\centering
\caption{Performance Comparison of Cocktail and different baseline KV cache quantization methods.}
\begin{tabular}{l|cccccccccc}
\toprule[1pt]
Model &
Method &
Qasper &
QMSum & 
MultiNews &
TREC &
TriviaQA &
SAMSum & 
LCC &
RepoBench-P &
Average
\\ 

\midrule
\multirow{5}{*}{\textbf{Llama2-7B}}
& FP16 & 9.63 & 21.32 & 3.47 & 66 & 87.74 & 41.81 & 66.62 & 59.77 & 44.55 \\
\cmidrule{2-11}
& Atom & 9.03 & 20.14 & 2.8 & 65.2 & 87.46 & 41.25 & 66.78 & 59.02 & 43.96\\
& KIVI & 9.18 & 20.66 & 1.38 & 65.7 & 87.42 & 41.53 & 66.45 & 59.43 & 43.97 \\
& KVQuant & 9.24 & 21.02 & 3.08 & 65.8 & 87.48 & 41.52 & 66.94 & 58.93 & 44.25\\
\rowcolor{orange!40}
\cellcolor{white}
& Cocktail & \textbf{9.67} & \textbf{21.21} & \textbf{3.21} & \textbf{66} & \textbf{87.66} & \textbf{42.08} & \textbf{67.58} & 59.1 & \textbf{44.56} \\
\midrule
\multirow{5}{*}{\textbf{Llama2-13B}}
& FP16 & 9.44 & 21.25 & 3.74 & 70.3 & 88.02 & 43.89 & 66.64 & 56.82 & 45.01 \\
\cmidrule{2-11}
& Atom & 8.63 & 20.34 & 4.56 & 69.4 & 86.75 & 43.18 & 64.22 & 55.35 & 44.05\\
& KIVI & 8.58 & 20.69 & 4.39 & 69.5 & 87.03 & 43.31 & 65.08 & 55.46 & 44.26\\
& KVQuant & 8.66 & 20.87 & 4.55 & 70 & 87.85 & 43.34 & 65.38 & 55.76 & 44.55\\
\rowcolor{orange!40}
\cellcolor{white}
& Cocktail & \textbf{8.79} & \textbf{21.03} & \textbf{4.82} & 70 & 87.37 & \textbf{43.97} & \textbf{66.15} & \textbf{56.65} & \textbf{44.85} \\
\midrule
\multirow{5}{*}{\textbf{Mistral-7B}}
& FP16 & 32.99 & 24.24 & 27.1 & 71 & 86.23 & 42.96 & 54.02 & 51.92 & 48.8 \\
\cmidrule{2-11}
& Atom & 30.54 & 23.66 & 26.35 & 70.32 & 85.79 & 42.18 & 53.27 & 51.64 & 47.97 \\
& KIVI & 30.72 & 23.98 & 26.93 & 67.8 & 86 & 43.34 & 53.59 & 51.73 & 48.39 \\
& KVQuant & 31.32 & 23.85 & 26.73 & 70.44 & 86.07 & 42.23 & 53.85 & 51.67 & 48.27 \\
\rowcolor{orange!40}
\cellcolor{white}
& Cocktail & \textbf{32.67} & \textbf{24.11} & \textbf{27.07} & \textbf{71} & \textbf{86.36} & \textbf{43.62} & 53.16 & \textbf{51.94} & \textbf{48.74}\\
\midrule
\multirow{5}{*}{\textbf{Longchat-7B}}
& FP16 & 29.41 & 22.77 & 26.61 & 66.5 & 83.99 & 40.9 & 52.94 & 56.78 & 47.49 \\
\cmidrule{2-11}
& Atom & 28.44 & 21.56 & 25.75 & 64.8 & 83.62 & 40.77 & 50.4 & 54.58 & 46.24\\
& KIVI & 28.69 & 22.59 & 26.28 & 66.5 & 83.19 & 40.28 & 52.4 & 55.13 & 46.88 \\
& KVQuant & 29.6 & 22.48 & 26.38 & 66.52 & 83.59 & 40.82 & 51.46 & 55.33 & 47.02\\
\rowcolor{orange!40}
\cellcolor{white}
& Cocktail & \textbf{31.88} & \textbf{22.65} & \textbf{26.43} & 66.5 & \textbf{83.88} & \textbf{41.1} & 51.95 & \textbf{55.42} & \textbf{47.48}\\
\bottomrule[1pt]

\end{tabular}
\label{tab:performance}
\vspace{-3mm}
\end{table*}
\begin{figure*}[htbp]
    \centering
    \begin{minipage}[b]{0.45\linewidth}
        \centering
        \includegraphics[width=1\linewidth]{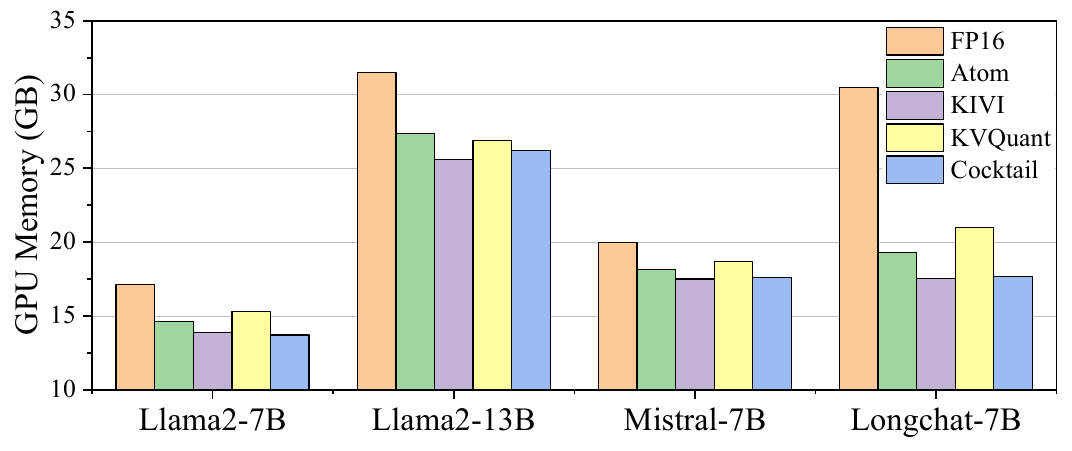}
        \caption{GPU memory of different models.}
        \label{fig:gpu_memory}
    \end{minipage}
    \hspace{0.05\linewidth}
    \begin{minipage}[b]{0.45\linewidth}
        \centering
        \includegraphics[width=1\linewidth]{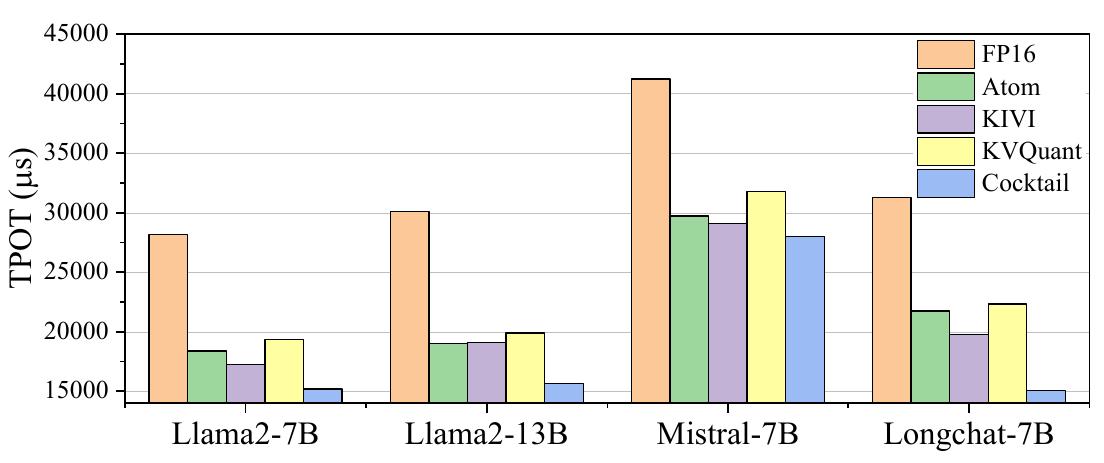}
        \caption{Time per output token (TPOT) of different models.}
        \label{fig:latency}
    \end{minipage}
    \vspace{-3mm}
\end{figure*}

\subsection{Experiment Setup}
\textbf{Models.} We evaluate Cocktail on four famous LLM models: Llama2-7B \cite{touvron2023llama}, Llama2-13B\cite{touvron2023llama}, Mistral-7B \cite{jiang2023mistral} and Longchat-7B \cite{li2023long}, where Longchat-7B is a specified fine-tuned model for chat tasks. The maximum context length of the first two models is 4K, while this number of the other two models is 32K. The output length is set as 128.

\textbf{Datasets.} We adopt LongBench\cite{bai2023longbench} benchmark for long context evaluation, the datasets in which are shown in Table 
\ref{tab:dataset}.

\textbf{Baselines.}
We only select three representative SOTA methods for comparison due to page limitations. They are Atom \cite{zhao2024atom}, representing trivial uniform quantization; KIVI \cite{liu2024kivi}, representing per-channel key quantization and per-token value quantization; and KVQuant \cite{hooper2024kvquant}, representing token-level mixed-precision quantization. The KV Cache is quantized to INT4 uniformly in both Atom and KIVI. As for KVQuant, a small portion (1\%) of the KV Cache retains FP16 precision, while the rest is quantized to INT4. Atom also includes functionality for quantizing weights and activations. However, to ensure a fair comparison, we do not use this functionality, focusing solely on the methods' ability to quantize the KV cache.

\textbf{Hardware.} We conduct all the experiments on an NVIDIA A800 GPU containing 80GB GPU memory, with a 25-core AMD EPYC 7T83 CPU and 100GB memory.

\subsection{Performance Comparison}
We compare the performance of Cocktail and the baseline methods on four models over eight different long-context datasets with $\alpha$ and $\beta$ set as 0.6 and 0.1, chunk size set as 32. The results are shown in Table \ref{tab:performance}. Cocktail achieves the best performance over most of the datasets, and the best average performance on all four models. Additionally, Cocktail has the least performance loss compared to the original FP16 precision model, with its performance score dropping by an average of only 0.055 compared to the original model. This is because we use chunk-level mixed-precision quantization to successfully maintain the precision of those important context parts.
\begin{figure}
    \centering
    \includegraphics[width=1\linewidth]{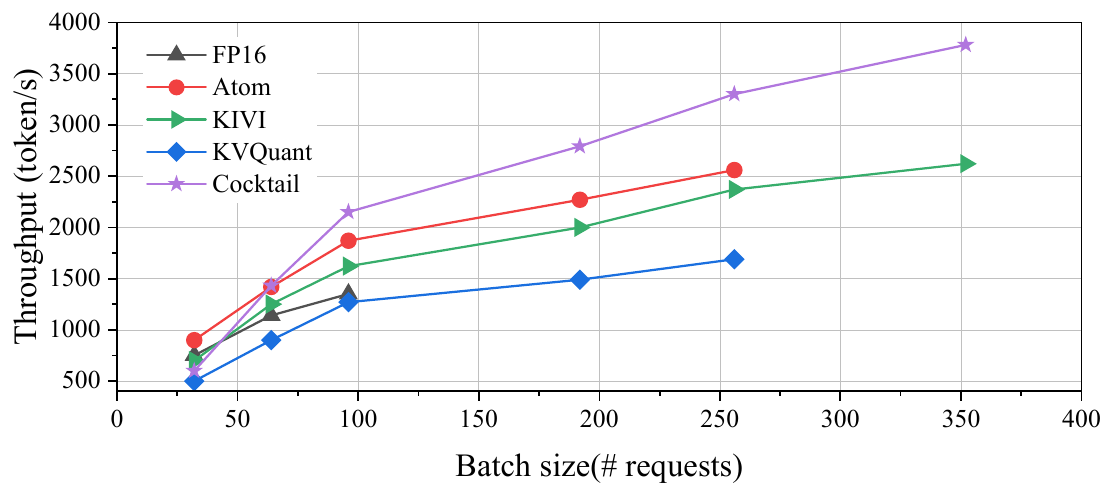}
    \caption{The throughput of different models with different batch sizes. The interrupted lines in the figure are due to OOM (Out of Memory).}
    \label{fig:throughput}
    \vspace{-5mm}
\end{figure}
\begin{table}[ht]
\centering
\caption{The impact of different chunk size on model performance.}
\label{tab:chunk size}
\begin{tabular}{c|c|c|c|c|c|c}
\toprule
Chunk Size   &                           8 & 16 & 32 & 64 & 128 & 256\\ \midrule
Rouge Score                             &  21.24 & 21.19  & 21.21  & 20.15 & 19.82 & 16.35       \\ \bottomrule
\end{tabular}
\vspace{-3mm}
\end{table}

Furthermore, we compare the GPU memory usage and time per output token (TPOT) of Cocktail and other methods on the QMSum dataset over four models. As shown in Figure \ref{fig:gpu_memory}, Cocktail can reduce the GPU memory usage by 12\%-42\% compared to the original FP16 precision model. Figure \ref{fig:latency} shows that Cocktail has the lowest TPOT, reducing by 32\%-52\% compared to the FP16 precision model. The improvements of Cocktail in GPU memory usage and TPOT are due to the use of chunk-level KV cache computation.

Besides, we evaluate the throughput of Cocktail and the baseline methods with different batch sizes on the QMSum dataset. As shown in Figure \ref{fig:throughput}, with the increase of batch size, the throughput of Cocktail is initially lower than the uniform quantization methods, but it gradually surpasses them. This is because when batch size is small, the latency of the chunk-level quantization search process limits the throughput of Cocktail. However, when the batch size is large, this latency is negligible compared with the LLM inference process. Since Cocktail's TPOT is smaller than that of the uniform quantization methods, its throughput is naturally higher. The throughput of Cocktail is always higher than KVQuant because our chunk-level quantization search is faster than its token-level quantization search.

\subsection{Analysis}
\label{subsec:analysis}
\begin{figure}
    \centering
    \includegraphics[width=0.48\textwidth]{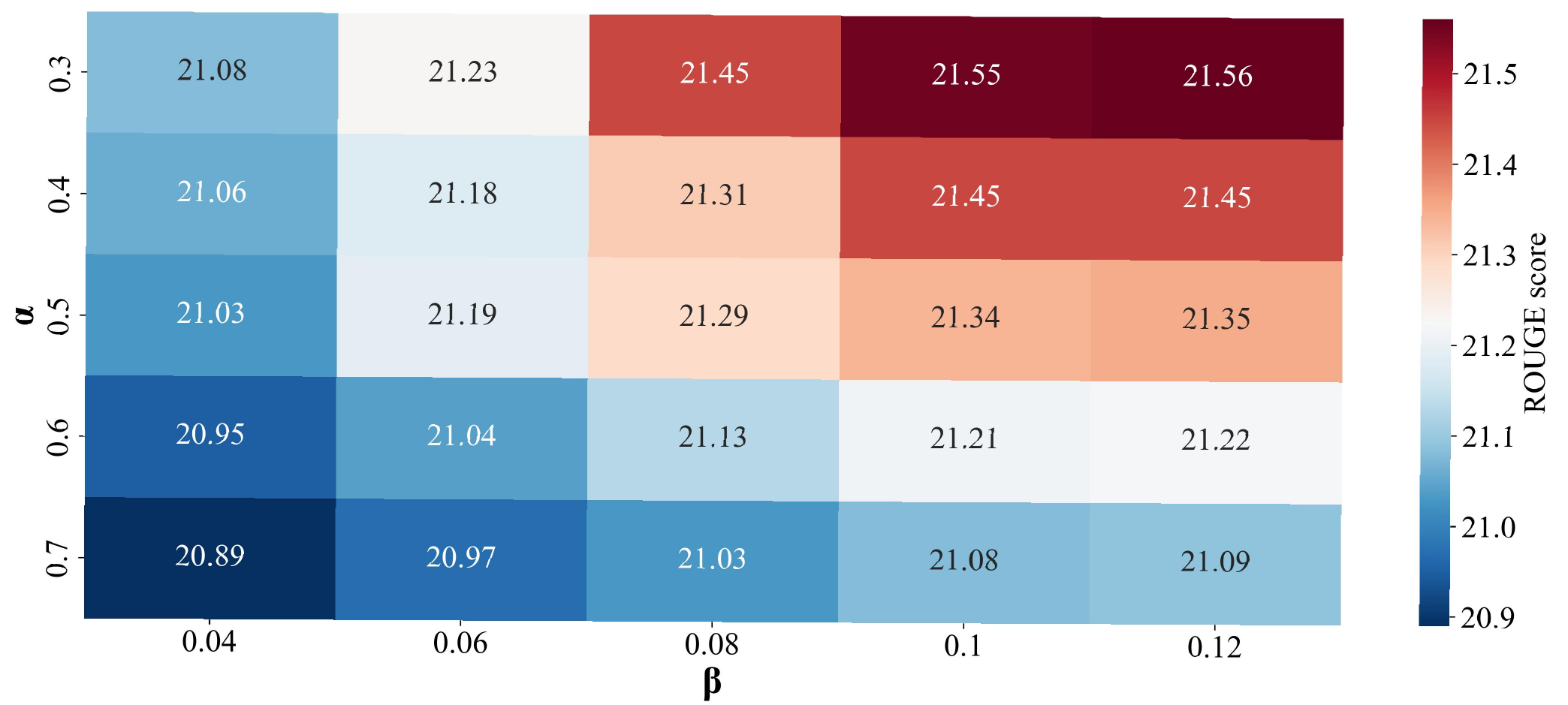}
    \caption{The impact of $\alpha$ and $\beta$ to model performance.}
    \label{fig:alpha_beta}
    \vspace{-2mm}
    
\end{figure}

We evaluate the impact of $\alpha$ and $\beta$ on the Llama2-7B model on QMSum dataset. As shown in Figure \ref{fig:alpha_beta},  the model's accuracy worsens as $\alpha$ increases and improves as $\beta$ increases. When $\beta$ increases to a certain extent, further changes in the model's accuracy become less noticeable. This is because a larger $\alpha$ means more context is quantized to INT2. Conversely, a larger $\beta$ means more context is retained in FP16 precision. However, when $\beta$ becomes sufficiently large, since only a few chunks of the context are closely related to the query, retaining more chunks in FP16 precision doesn't contribute much to improving model accuracy. We also evaluate the impact of different chunk sizes. As shown in Table \ref{tab:chunk size}, when the chunk size is smaller than 32, the model performance stays steady. But when the chunk size is larger than 32, the model accuracy drops quickly. This is because the important context parts within a chunk of large size can be surrounded by many unimportant parts, leading to incorrect quantization to lower bits.

Furthermore, we explore different context and query encoder architectures in the chunk-level quantization search module. The results are shown in Table \ref{tab:encoder}. We select four prevalent encoder: ADA-002 \cite{ADA-002}, BM25 \cite{lin2021pyserini}, LLM Embedder \cite{wu2022memorizing}, and Facebook-Contriever \cite{izacard2021unsupervised}. Over four different datasets, the Facebook-Contriever encoder has the best performance. Therefore, we choose it as our context and query encoder.
\subsection{Ablation Study}
We conduct ablation studies to prove the effect of the two modules in Cocktail. The experiments are conducted on the QMSum dataset on Llama2-7B model, with results shown in Table \ref{tab:ablation} (The two modules are referred to as module \uppercase\expandafter{\romannumeral1} and \uppercase\expandafter{\romannumeral2}, respectively). Without chunk-level quantization search, the model accuracy can drop drastically. On the other hand, without chunk-level KV cache computation, GPU memory usage and TPOT can increase significantly. This experiment illustrates the effect of chunk-level mixed-precision quantization on maintaining model accuracy and the effect of chunk-level KV cache computation on reducing GPU memory usage and latency.

\begin{table}[htbp]

\centering
\caption{Performance Comparison of Different context encoder and query encoder on LLama2-7B over four datasets.}
\label{tab:encoder}
\resizebox{\columnwidth}{!}{%
\begin{tabular}{c|c|c|c|c}
\toprule
Method                               & Qasper & SAMSum & TriviaQA & RepoBench-P \\ \midrule
Baseline                             & 9.52   & 41.69  & 87.72    & 59.82       \\ \midrule
ADA-002 \cite{ADA-002}               & 8.89   & 41.34  & 86.69    & 58.32       \\
BM25 \cite{lin2021pyserini}          & 7.36   & 35.67  & 84.22    & 55.35       \\
LLM Embedder \cite{wu2022memorizing} & 8.98   & 41.56  & 87.85    & 59.77       \\

Facebook-Contriever \cite{izacard2021unsupervised} & 9.67 & 42.08 & 88.06 & 60.1 \\ \bottomrule
\end{tabular}%
}
\vspace{-2mm}
\end{table}
\begin{table}[htbp]
\vspace{-3mm}
\caption{The impact of chunk-level quantization search and
Chunk-level KV cache computation.}
    \centering
    \begin{tabular}{c|c|c|c}
    \toprule
    Method & F1 score $\uparrow$ & GPU Memory(GB) $\downarrow$ & TPOT($\mu$s) $\downarrow$ \\ 
    \midrule
    Baseline & 21.32 & 17.13 & 28214  \\
    \midrule
    w/o Module \uppercase\expandafter{\romannumeral1} & 19.33 & 13.88 &  15258 \\
    w/o Module \uppercase\expandafter{\romannumeral2} & 21.21 & 20.6 & 29653 \\
    Cocktail & 21.21 & 13.7 & 15201 \\
    \bottomrule
    \end{tabular}
    
    \label{tab:ablation}
    
\end{table}

\section{Conclusion}
This paper introduces Cocktail, a chunk-adaptive mixed-precision KV cache quantization method for long-context LLM inference. Cocktail uses a chunk-level quantization search module to determine the bitwidth configuration of context KV cache chunks. It also contains a chunk-level KV cache computation module, reordering these KV cache chunks to avoid hardware inefficiency. Extensive experiments on multiple datasets and models demonstrate that Cocktail outperforms state-of-the-art KV cache quantization methods.

\section{Acknowledgements}
This work was sponsored by the Key Research and Development Program of Guangdong Province under grant No. 2021B0101400003, the National Key Research and Development Program of China under Grant No.2023YFB4502701.

\newpage

\bibliographystyle{unsrt}
\bibliography{date}
\end{document}